\documentclass[10pt,twocolumn,letterpaper]{article}

\usepackage{cvpr}              % To produce the CAMERA-READY version
\usepackage{multirow}
\usepackage{makecell} % 导言区
\usepackage{url}

\definecolor{cvprblue}{rgb}{0.21,0.49,0.74}
\usepackage[pagebackref,breaklinks,colorlinks,allcolors=cvprblue]{hyperref}
\usepackage{url}

\def\paperID{30900} % *** Enter the Paper ID here
\def\confName{CVPR}
\def\confYear{2026}

\title{DIMOS: Disentangling Instance-level Moving Object Segmentation}

\author{
Hongxiang Huang\quad
Hongwei Ren\quad
Xiaopeng Lin\quad
Yulong Huang\quad
Zeke Xie\quad
Bojun Cheng\\
The Hong Kong University of Science and Technology (Guangzhou)\\
Guangzhou, China\\
{\tt\small hhuang516@connect.hkust-gz.edu.cn \quad bocheng@hkust-gz.edu.cn}
}

\begin{document}
\maketitle
\begin{abstract}
Moving instance segmentation (MIS) attracts increasing attention due to its broad applications in traffic surveillance, autonomous driving, and animal tracking. Event cameras record asynchronous brightness changes, providing high temporal resolution and dynamic range, which makes them highly sensitive to motion information. By fusing event and image features, motion cues from events can complement spatial details from images, enhancing the performance of MIS. However, current multimodal MIS methods still struggle to segment small moving instances, as event cameras often yield sparse features under limited resolution. In addition, event features entangle appearance attributes with motion cues, which further restricts effective cross-modal fusion. To address these challenges, we first propose a dual-disentangling feature extraction framework that separates and extracts appearance and motion information within both image and event modalities, thereby improving feature density. Subsequently, a multi-granularity cross-modal alignment is introduced to align distributionally and semantically consistent features across modalities, enabling more effective fusion with rich spatial and temporal details. The experiment results demonstrate that our method achieves state-of-the-art performance in multimodal MIS, especially for small instances under challenging conditions such as fast motion and low-light settings.
% \footnote{Project page: \url{https://github.com/Neuromorphic-Electronics-Photonics-Lab/DIMOS-Moving-Instance-Segmentation-CVPR2026}.}
\end{abstract}    
\section{Introduction}
\label{sec:intro}

\begin{figure}[t]
  \centering
   \includegraphics[width=1.0\linewidth]{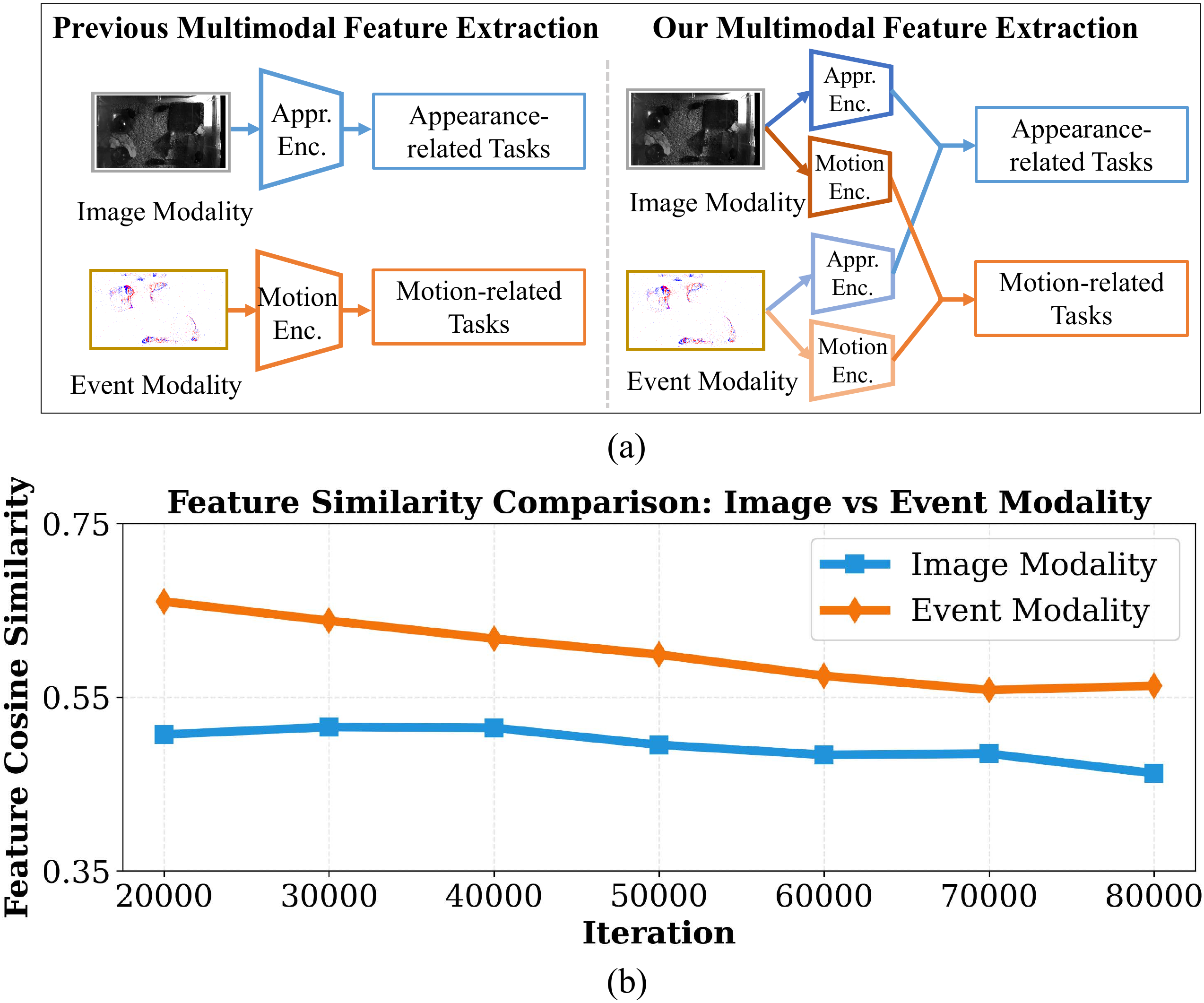}

  \caption{Feature Extraction Comparison.
(a) Our method extracts both appearance and motion features from each modality.
(b) Model checkpoints at different iterations are sampled to compute the cosine similarity between appearance and motion features extracted from encoders of the same modality on MouseSIS dataset. 
%Compared to models without disentanglement, the disentangling models exhibit a more pronounced decrease in feature similarity during training.
}
\label{fig:feature_comparison}
\end{figure}

Moving instance segmentation has gained increasing attention owing to its wide applications in traffic surveillance~\cite{zhang2020traffic}, autonomous driving~\cite{zeller2023radar}, and animal tracking~\cite{hamann2024mousesis}. This task is inherently more challenging than conventional semantic segmentation, as it requires not only distinguishing object categories but also segmenting individual instances and identifying their independent motions. Recent advances in moving instance segmentation algorithms have achieved impressive results~\cite{wu2022defense, cheng2022xmem}. However, the performance of image-based methods still shows limitations in extreme environments, such as low illumination, backlighting, and high-speed motion scenarios~\cite{hamann2024mousesis, li2024event}. 

Event cameras, with their unique advantages in low latency, high dynamic range, and low power consumption, have shown potential to overcome the limitations of image-based methods~\cite{fu2026spikeram,huang2024clif}. Built on more complex pixel circuits, event cameras achieve extremely high temporal resolution but relatively low spatial resolution~\cite{lichtsteiner2008128}. In addition, they generate sparse and asynchronous events rather than dense frame-based signals. Although these characteristics are advantageous for capturing rapid motion in extreme environments~\cite{yao2021temporal,lin2026event,ren2025e2b,ren2025rethinking,huang2025exploring}, they also introduce challenges for dense prediction tasks such as instance segmentation. In particular, the sparsity of event streams leads to incomplete spatial context, making it difficult to capture clear object boundaries and regions. As a result, although event data are valuable for mitigating limitations under extreme conditions, models relying solely on event inputs still struggle to achieve competitive performance in moving instance segmentation. 

Recent research has increasingly explored multimodal fusion~\cite{li2025efficient, xie2024eisnet,lin2025clearsight}, in which each modality provides complementary strengths. Image frames provide appearance cues such as texture and structural details, while event streams provide motion information. This paradigm shows clear performance improvements. However, event cameras still suffer from limited resolution due to their larger pixel pitch, and current fusion methods often require maintaining consistent resolution between image and event sensors~\cite{wan2025instance, hamann2025sis}. These factors make small instance segmentation very challenging. Since small objects occupy only a limited number of pixels, both appearance and motion information become constrained. This sparsity leads to insufficient feature density and degraded segmentation quality. In practice, each modality naturally contains both types of cues. Image frames contain motion cues, as widely utilized in optical flow estimation~\cite{dosovitskiy2015flownet, ranjan2017optical}. The density or distribution of event streams provides implicit clues about appearance patterns related to shape, texture, and material, as these attributes determine the surface reflectance~\cite{yu2025eventpsr, mitrokhin2020learning, gallego2018unifying, gallego2020event}. Such inherent properties have not been fully explored or exploited in moving instance segmentation.

A promising solution is to extract both appearance and motion cues from each modality instead of relying on a strict separation between them, which increases feature density and enables more effective use of the available pixels, as shown in Figure~\ref{fig:feature_comparison}(a). However, achieving such joint extraction is not equally straightforward for different modalities. For image data, the separation between appearance and motion cues is well-understood. Appearance details are naturally captured by the camera, while motion information can be derived from temporal differences or motion blur~\cite{gong2017motion}. In contrast, event data exhibits strong entanglement between appearance and motion cues because both motion and appearance characteristics can induce variations in event density and distribution that are difficult to distinguish. This coupling complicates the extraction of clean appearance and motion features from the event modality and ultimately weakens the effectiveness of cross-modal fusion. As shown in Figure~\ref{fig:feature_comparison}(b), features extracted from the event modality exhibit higher similarity across different types compared to those from the image modality. To solve the entanglement problem, feature disentanglement has been explored in other dense prediction tasks, such as image segmentation~\cite{wei2023disentangle, pei2021disentangle}, super resolution~\cite{wang2022disentangling, kong2025dam}, and image generation~\cite{dai2023disentangling,dai2025beyond}, where it helps isolate different semantic factors and improve feature interpretability. These observations motivate an intra-modal disentanglement strategy to effectively separate appearance and motion cues, especially for event data.

In this work, we focus on disentangling and extracting both appearance and motion features within each modality. Specifically, we design a Disentangling Instance-level Moving Object Segmentation (DIMOS) framework, using dual-disentangling encoders with intra-modal contrastive learning and task-specific supervision to disentangle appearance and motion features from both image and event modalities. Intra-modal contrastive learning enhances the discriminability between appearance and motion information, while task-specific supervision constrains the disentangled features to learn appearance and motion cues, respectively. Since disentanglement produces two types of features per modality, effective multimodal learning requires aligning appearance and motion features across modalities. To this end, we introduce a multi-granularity cross-modal alignment mechanism that combines adversarial domain adaptation and modality translation. This design facilitates effective feature fusion by jointly enforcing distributional and semantic alignment between cross-modal features.
\begin{itemize}
  
\item We propose a dual-disentangling mechanism that extracts both appearance and motion features from each modality, enhancing feature density and representation quality.

\item We design a multi-granularity cross-modal alignment to enforce distributional and semantic consistency for effective feature fusion.

\item Experiments demonstrate that our approach achieves state-of-the-art performance, validating the effectiveness of the proposed disentanglement framework.

\end{itemize}
\section{Related Work}
\label{sec:related}

\subsection{Video Object Segmentation}

Video object segmentation (VOS) lays the foundation for most moving object and instance segmentation tasks. In the semi-supervised setting, an initial annotation in the first frame is propagated over time. Early methods focused on efficient temporal propagation, while later methods emphasized robust feature matching. Representative methods such as FEELVOS~\cite{voigtlaender2019feelvos}, CFBI~\cite{yang2020collaborative}, and CFBI+~\cite{yang2021collaborative} significantly improved temporal consistency and stability, forming the basis for many subsequent frameworks.

Beyond, unsupervised or zero-shot VOS methods further relax need for an initial annotation by relying only on the intrinsic visual and motion cues in videos. MATNet~\cite{zhou2020matnet} combines appearance and motion for foreground discovery, while Isomer~\cite{yuan2023isomer} leverages transformer-based architectures for long-range temporal modeling. Another approach~\cite{zhao2024adaptive} explores multi-source fusion to improve segmentation accuracy in more diverse scenarios. Although these approaches improve generalization, they remain sensitive to complex camera motion and adverse conditions.
To address this limitation, recent works integrate event data into VOS. Event cameras offer high temporal resolution and clean motion cues under challenging lighting conditions. ELVOS~\cite{li2024event} demonstrates that fusing event streams with images significantly improves temporal correspondence and segmentation reliability, highlighting cross-modal fusion as a promising strategy for robust VOS.

\subsection{Moving Instance Segmentation}

Moving instance segmentation (MIS) extends VOS from foreground localization to distinguishing multiple independently moving objects. 
Early motion segmentation approaches, such as FgSegNet~\cite{ang2018foreground}, primarily focus on identifying moving regions through background subtraction, optical flow estimation, or motion clustering. 
Event-based variants, including EVIMO~\cite{mitrokhin2019ev}, GConv~\cite{mitrokhin2020learning}, and Un-EVIMO~\cite{wang2024evimo}, further leverage the high temporal resolution of event cameras to separate independent motions. 
However, due to the sparse and asynchronous nature of event data, these approaches often produce coarse object boundaries and incomplete contours, limiting their performance in fine-grained instance-level segmentation.
To achieve more precise instance discrimination, image-based MIS methods adapt video instance segmentation (VIS) architectures that exploit rich appearance cues. 
IDOL~\cite{wu2022defense} introduces an interaction mechanism between detection and segmentation to maintain temporal consistency. 
Although it performs well in structured scenes, its reliance on single-modality inputs restricts robustness under motion blur, occlusion, and other degraded conditions.

To overcome these limitations, multimodal pipelines emerge to integrate complementary cues from both images and events. 
ModelMixSort~\cite{hamann2024mousesis} combines YOLO~\cite{redmon2016you} detectors on RGB and event-derived grayscale frames with a SAM~\cite{kirillov2023segment, dai2025vg} model, whereas EvInsMOS~\cite{wan2025instance} explicitly fuses image texture and event-based motion cues through contrastive learning~\cite{oord2018representation,dai2024one} and cross-modal masked attention. 
Despite these advances, most multimodal MIS frameworks still follow a simplified paradigm that extracts appearance information from images and motion information from events. 
Such designs often result in insufficient feature density for small objects and weak semantic correspondence across modalities, highlighting the necessity for a unified framework that jointly disentangles and aligns appearance–motion representations to achieve more robust moving instance segmentation.

\section{Method}
\label{sec:method}

\begin{figure*}[t]
  \centering
   \includegraphics[width=0.95\linewidth]{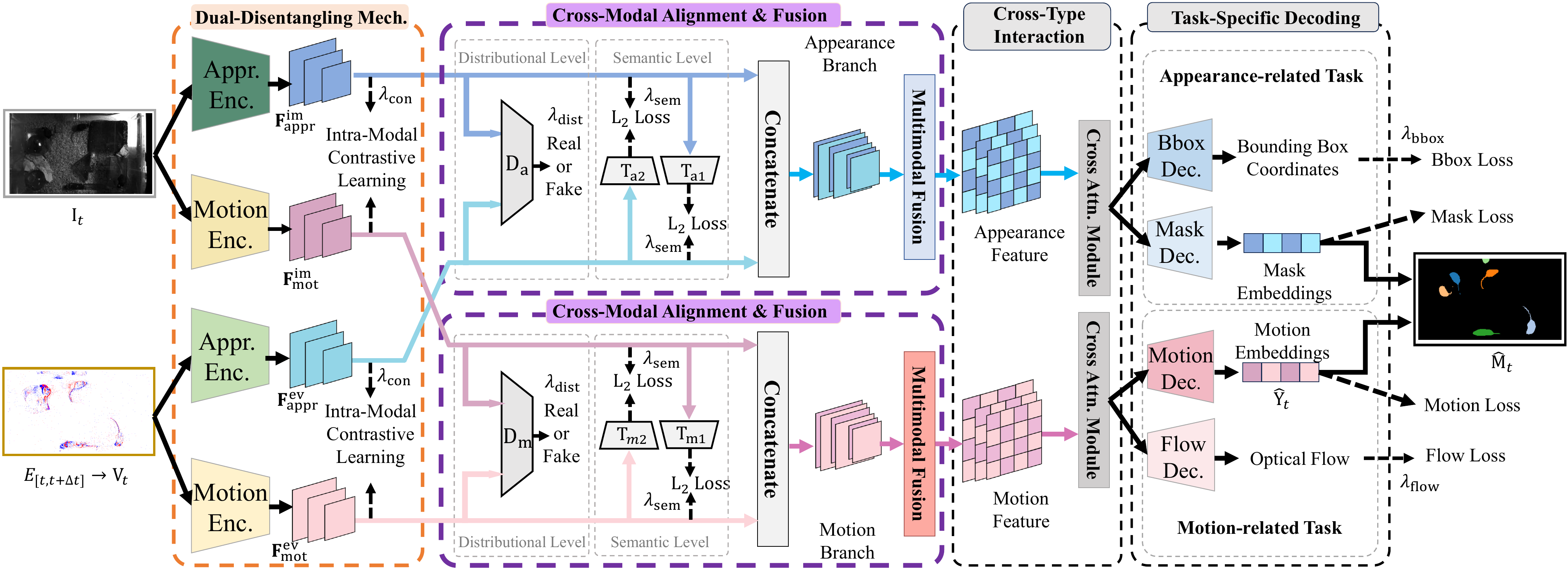}

   \caption{
\textbf{Overview of the proposed DIMOS framework.} 
The pipeline consists of four major components: (1) \textbf{Dual-Disentangling Mechanism} with appearance and motion encoders for each modality. (2) \textbf{Multi-Granularity Cross-Modal Alignment \& Fusion} that enforces consistency at both distributional and semantical levels. (3) \textbf{Cross-Type Interaction} for joint reasoning between appearance and motion cues via cross attention. and (4) \textbf{Task-Specific Decoders} for appearance-related and motion-related predictions. 
$D_a$ and $D_m$ denote the domain discriminators for appearance and motion branches used in adversarial distribution alignment, while $T_{a1}$, $T_{a2}$, $T_{m1}$, and $T_{m2}$ represent modality translation modules that perform bidirectional feature reconstruction for semantic-level alignment.
}

   \label{fig:overview}
\end{figure*}

In this section, we present the \textbf{Disentangling Instance-Level Moving Object Segmentation (DIMOS)} framework. It extracts both appearance and motion features from each modality and introduces feature disentanglement with multi-granularity cross-modal alignment to enhance feature quality and fusion robustness.

% Our overall method builds upon a classical encoder–decoder architecture, but incorporates systematic modifications specifically designed for multimodal disentanglement and fusion. 
% The core ideas are summarized as follows:

% \begin{itemize}
%     \item \textbf{Disentangled Feature Extraction (Sec.~\ref{sec:disentangle})}: 
%     We design a dual-disentangling architecture that extracts appearance and motion features in parallel within each modality. These branches are jointly optimized using contrastive learning and task-specific supervision, which enhances both feature discriminability and disentanglement quality.
%     \item \textbf{Cross-Modal Alignment (Sec.~\ref{sec:alignment})}: 
%     We propose a multi-granularity alignment mechanism to align features across modalities at both distributional and semantic levels. Adversarial domain adaptation ensures coarse-grained distribution alignment, while a modality translation module enforces fine-grained semantic consistency.
%     \item \textbf{Joint Optimization (Sec.~\ref{sec:optimization})}: 
%     We integrate multiple supervision signals—including instance segmentation, motion prediction, detection, contrastive disentanglement, and cross-modal alignment—into an end-to-end training objective.
% \end{itemize}

% As illustrated in Fig.~\ref{fig:framework}, our DIMOS framework substantially enhances segmentation performance for small objects under limited-resolution settings, while maintaining a lightweight and flexible structure.

\subsection{Problem Definition}
\label{sec:problem}

Given a sequence of image frames $\mathbf{I}_t$ and the corresponding event stream $\mathbf{E}_{[t, t+\Delta t]}$ recorded by an event camera over the same time interval, our goal is to perform pixel-level segmentation of all independently moving objects in the scene and to predict their motion states at the current time step. 
Specifically, the model is required to predict an instance mask $\hat{m}_k$ and a binary motion label $\hat{y}_k \in \{0, 1\}$ for each instance. 
The overall output can be expressed as follows,
\begin{equation}
    \hat{\mathbf{M}} = \{\hat{m}_k\}_{k=1}^K, \quad 
    \hat{\mathbf{Y}} = \{\hat{y}_k\}_{k=1}^K,
\end{equation}
where $K$ denotes the number of instances in the current frame.

Event cameras produce asynchronous event streams 
$\mathbf{E} = \{e_i\}_{i=1}^N$, 
where each event $e_i = (x_i, y_i, t_i, p_i)$ encodes a brightness change at time $t_i$ and pixel $(x_i, y_i)$ with polarity $p_i$.
While event data offers ultra-high temporal resolution, it is inherently sparse and irregular, making it less suitable for dense prediction tasks.
To address this, we discretize the event stream into $B$ temporal bins and accumulate events spatially and temporally to obtain a voxel representation as
\begin{equation}
\mathbf{V}_t(x, y, b)
= \sum_{\substack{i \\ (x_i, y_i) = (x, y)}} 
p_i \,
\max\Bigl(
0,\,
1 - \bigl| b - \tfrac{t_i - t}{\Delta t}(B - 1) \bigr|
\Bigr),
\end{equation}
where $b \in \{0,1,\dots,B-1\}$ is the bin index, $B$ is the total number of bins, and $\Delta t$ denotes the duration of the event slice between the current frame and the next frame. 

The two modalities are combined as $\mathbf{X} = \{\mathbf{I}_t, \mathbf{V_t}\}$ and assigned to the segmentation output through a mapping function with learnable parameters $\phi$ as follows,
\begin{equation}
    f_\phi(\mathbf{X}) = (\hat{\mathbf{M}}, \hat{\mathbf{Y}}).
\end{equation}

\subsection{Method Overview}
%Our framework follows the paradigm with multi-scale encoding and query-based decoding commonly used in VIS or VOS~\cite{cheng2022mask2former, yang2021cfbi}.  However, we introduce several key innovations.  Both image and event inputs are independently encoded by dual-branch encoders to extract appearance and motion features, where each branch adopts a standard pretrained CNN backbone. Both dual-branch outputs are then refined through intra-modal contrastive learning to enhance disentanglement as shown inside the orange dashed boxes in Figure~\ref{fig:overview}.Aligned appearance and motion features from image and event modalities are fused through the multi-granularity cross-modal alignment and a lightweight CNN fusion layer as illustrated inside the purple dashed boxes in Figure~\ref{fig:overview}. 
 
%Afterward, two widely used cross-attention modules~\cite{} are adopted to model correlations between appearance and motion cues. On top of the fused representations, we employ parallel decoders for appearance-related (instance segmentation and bounding box regression) and motion-related (motion classification and optical flow estimation) tasks, which ensure the effective semantic separation during the disentanglement.  

Our framework consists of four main components: a dual-disentangling module, a cross-modal alignment and fusion module, a cross-type interaction module, and a task-specific module. Compared to previous works in VIS~\cite{cheng2022masked, wu2022seqformer}, we make several key modifications. First, both image and event are encoded independently with a dual-branch encoder to extract appearance and motion features, as shown inside the orange dashed box in Figure~\ref{fig:overview}. To ensure a clear separation of appearance and motion features, we adopt intra-modal contrastive learning to enhance disentanglement within each branch. Second, the two inputs yield four feature vectors that are subsequently fused across the image and event modalities into appearance and motion features. This is accomplished by a multi-granularity cross-modal alignment module and a lightweight CNN fusion layer for the image and event modalities, as shown inside the purple dashed boxes in Figure~\ref{fig:overview}. Third, to preserve the semantics of disentangled features, we employ not only instance segmentation and bounding box regression for decoders of appearance-related tasks, but also motion classification and optical flow estimation for decoders of motion-related tasks. 

We employ different strategies for inference and training.
During inference, we utilize the mask fusion procedure proposed in~\cite{wan2025instance}. The model first upsamples the predicted mask embeddings to full resolution and then fuses the motion classification results by applying a confidence threshold $\theta$ to the motion scores. Only masks whose confidence exceeds $\theta$ are retained as moving instances, producing the final instance-level segmentation map. In contrast, during training, the supervision is assigned by Hungarian matching between predicted masks and ground-truth instances~\cite{cheng2021per}, ensuring a one-to-one correspondence for both mask prediction and motion classification without thresholding.

\subsection{Dual-Disentangling Mechanism}
\label{sec:dual_disentangling}

%Small objects occupy only a very limited number of pixels, making effective pixel utilization critical in low-resolution scenarios. If appearance features are only extracted from images and motion features only from events, the overall feature density becomes insufficient, leading to degraded segmentation performance.

%To obtain sufficient information, we aim to extract both appearance and motion information from each modality and leverage task biases to strengthen these representations. Image frames naturally encode appearance texture and structure, but when motion blur occurs or when consecutive frames are compared, they also implicitly carry motion signals. Conversely, while event streams are triggered by changes in brightness, their spatial distribution and density patterns are closely related to an object’s shape, texture, and structure, which means they also contain appearance attributes.
%However, unlike the image modality where appearance and motion features are relatively explicit, event data exhibit strong coupling between appearance and motion signals due to their motion-triggered nature. For instance, event edges not only indicate an object’s motion trajectory but are also shaped by its material, texture, and geometry. This coupling makes direct disentanglement of event features more challenging. Consequently, effectively separating appearance and motion representations within each modality becomes a core motivation of our architectural design.

To ensure sufficient information for small instance segmentation, we aim to jointly exploit appearance and motion information extracted from each modality. Providing complementary information from both the image and the event modalities can alleviate the lack of dense information within a single modality. Therefore, we adopt a dual-branch encoder for each modality to simultaneously extract appearance and motion information from image and event inputs. 

The dual-branch encoders of the two modalities share the same input but are parameterized independently and trained with different task-specific supervision signals to learn distinct semantic representations.
% To address this, we propose a \textbf{dual-disentangling architecture}. 
% In the encoder of each modality, we introduce two parallel branches: one branch focuses on extracting appearance-biased features, while the other captures motion-biased features. 
% These two branches share the same input but are parameterized independently and are trained under different task-specific supervision signals that encourage them to learn semantically distinct representations. 
We denote the appearance and motion features extracted from images as $\mathbf{F}^{\mathrm{im}}_{\mathrm{appr}}$ and $\mathbf{F}^{\mathrm{im}}_{\mathrm{mot}}$, 
and those from events as $\mathbf{F}^{\mathrm{ev}}_{\mathrm{appr}}$ and $\mathbf{F}^{\mathrm{ev}}_{\mathrm{mot}}$. 
To further enhance the disentanglement between the appearance and motion branches, we incorporate \textbf{intra-modal contrastive learning}. Unlike previous works that apply contrastive learning for improving cross-modal feature discriminability, we focus on intra-modal separation of appearance and motion features. This encourages the network to emphasize differences between appearance and motion semantics rather than across modalities, avoiding redundant or mixed representations across branches.
For each modality, positive samples $\mathbf{F}^+$ are selected from the same type (appearance or motion) and consecutive frames, while negative samples $\mathbf{F}^-$ are sampled from different types or non-consecutive frames. The InfoNCE loss~\cite{oord2018representation} for intra-modal contrastive learning is defined as
\begin{equation}
    \mathcal{L}_{\mathrm{con}} 
    = - \log 
    \frac{\exp\big(\mathbf{F} \cdot \mathbf{F}^+ / \tau\big)}
    {\exp\big(\mathbf{F} \cdot \mathbf{F}^+ / \tau\big) + 
    \sum_{\mathbf{F}^-} \exp\big(\mathbf{F} \cdot \mathbf{F}^- / \tau\big)},
\end{equation}
where $\cdot$ denotes dot product between two $\ell_2$-normalized features and $\tau$ is a temperature factor.
The specific construction of positive and negative samples for appearance and motion features is detailed in the supplementary material.

\subsection{Multi-Granularity Cross-Modal Alignment}
\label{sec:cross_modal_alignment}
 
Previous multimodal fusion methods often combine features from different modalities through concatenation and simple linear or convolutional operations. Without proper alignment before fusion, these methods fail to ensure semantic consistency or fully exploit the complementarity between image and event data. This limitation arises from the inherent distributional and semantic gap between the two modalities.  
Therefore, we propose a \textbf{multi-granularity cross-modal alignment} that aligns appearance and motion features across modalities by enforcing consistency at both distributional and semantic levels during the feature fusion.

\textbf{Distribution-Level Alignment via Domain Adaptation.}
%Although image and event data originate from the same scene, their feature statistics differ significantly: image features emphasize dense spatial textures, while event features arise from sparse temporal triggers. 
At the distributional level, the two modalities can be regarded as two “domains” of the same underlying scene. The distribution gap between them makes feature alignment difficult, even if they share the same semantic category (e.g., appearance or motion).
To bridge this gap, we employ \textbf{adversarial domain adaptation}~\cite{tzeng2017adversarial} to learn domain-invariant representations.
%, encouraging the encoder to fool a domain discriminator.

We introduce two discriminators for the appearance and motion branches. The discriminators classify feature modality, while the encoders, through a gradient reversal layer, learn to minimize this gap. 
An asymmetric strategy is adopted, where image features serve as the reference domain for appearance alignment and event features for motion alignment. This design leverages the fact that images provide more explicit appearance cues, while events contain clearer motion cues.
The adversarial loss is as follows:
\begin{equation}
    \mathcal{L}_{\text{adv}} = \mathcal{L}_{\text{adv}}^{\text{appr}} + \mathcal{L}_{\text{adv}}^{\text{mot}},
\end{equation}
where each component corresponds to an adversarial training objective between the encoder and its corresponding domain discriminator. The adversarial loss can be formulated as a min–max optimization as below,
\begin{equation}
    \min_{G} \max_{D} \; 
    \mathbb{E}_{x \sim p_{\text{ref}}} \big[\log D(x)\big] +
    \mathbb{E}_{z \sim p_{\text{src}}} \big[\log (1 - D(G(z)))\big],
\end{equation}
where $G$ denotes the feature encoder, $D$ the domain discriminator, $p_{\text{ref}}$ the reference domain distribution, and $p_{\text{src}}$ the source domain distribution.
For the appearance branch, $x=\mathbf{F}^{\mathrm{im}}_{\mathrm{appr}}$ and $G(z)=\mathbf{F}^{\mathrm{ev}}_{\mathrm{appr}}$.
For the motion branch, $x=\mathbf{F}^{\mathrm{ev}}_{\mathrm{mot}}$ and $G(z)=\mathbf{F}^{\mathrm{im}}_{\mathrm{mot}}$.
Detailed formulations are provided in the supplementary material.
% For the \textbf{appearance branch}, $x$ corresponds to $\mathbf{F}^{\mathrm{im}}_{\mathrm{appr}}$ for the reference domain and $G(z)$ corresponds to $\mathbf{F}^{\mathrm{ev}}_{\mathrm{appr}}$ for the source domain.  
% For the \textbf{motion branch}, $x$ corresponds to $\mathbf{F}^{\mathrm{ev}}_{\mathrm{mot}}$ for the reference domain and $G(z)$ corresponds to $\mathbf{F}^{\mathrm{im}}_{\mathrm{mot}}$ for the source domain. The detailed formulation of the distribution-level alignment loss for appearance and motion branches is provided in the supplementary material.

\textbf{Semantic-Level Alignment via Modality Translation.}
Distribution-level alignment alone cannot fully ensure semantic consistency across modalities. To bridge this gap, we introduce two lightweight convolutional \textbf{modality transform modules} that translate appearance and motion features between the image and event spaces, enforcing bidirectional semantic consistency during training. 
A reconstruction loss regularizes this process as below,
\begin{equation}
\begin{aligned}
    \mathcal{L}_{\text{trans}} = &
    \| T_{a1}(\mathbf{F}^{\mathrm{im}}_{\mathrm{appr}}) - \mathbf{F}^{\mathrm{ev}}_{\mathrm{appr}} \|_2^2 
  + \| T_{a2}(\mathbf{F}^{\mathrm{ev}}_{\mathrm{appr}}) - \mathbf{F}^{\mathrm{im}}_{\mathrm{appr}} \|_2^2 \\
  &+ \| T_{m1}(\mathbf{F}^{\mathrm{im}}_{\mathrm{mot}}) - \mathbf{F}^{\mathrm{ev}}_{\mathrm{mot}} \|_2^2
  + \| T_{m2}(\mathbf{F}^{\mathrm{ev}}_{\mathrm{mot}}) - \mathbf{F}^{\mathrm{im}}_{\mathrm{mot}} \|_2^2,
\end{aligned}
\end{equation}
where $T$ is the translation module.
This design ensures that features of the same semantic type are mutually translatable, strengthening cross-modal alignment and providing a more stable foundation for fusion. 

Importantly, both distributional and semantic alignments are entirely unsupervised and only applied during training, introducing no extra cost at inference time.

\subsection{Optimizing Objectives}
\label{sec:optimizing_objectives}

The overall training objective of DIMOS integrates task-specific supervision, intra-modal contrastive learning, and cross-modal alignment.

\textbf{Main Task Loss.}  
Following Sec.~\ref{sec:problem}, the model predicts both instance masks and motion states.  
The instance segmentation loss is defined as
\begin{equation}
    \mathcal{L}_{\mathrm{mov\_seg}} 
    = \frac{1}{K} \sum_{k=1}^{K} 
    \big[\mathcal{L}_{\mathrm{cls}}(\hat{y}_k, y_k) + 
    \mathcal{L}_{\mathrm{mask}}(\hat{m}_k, m_k)\big],
\end{equation}
where $y_k$ and $\hat{y}_k$ are the ground truth and predicted motion labels, and $m_k$ and $\hat{m}_k$ denote the ground truth and predicted masks. $\mathcal{L}_{\mathrm{cls}}$ is a standard cross-entropy loss used for class prediction, and $\mathcal{L}_{\mathrm{mask}}$ is a binary cross-entropy loss for mask supervision.

\textbf{Extra Task-Specific Loss.}  
To enhance appearance and motion perception, we introduce two extra objectives. For motion modeling, an unsupervised optical flow estimation loss~\cite{wan2025instance} is defined as
\begin{equation}
\mathcal{L}_{\mathrm{flow}}
= \sum_{\mathbf{c}} \psi \!\left(
\mathbf{I}_t(\mathbf{c}) - \mathbf{I}_{t+\Delta}\!\big(\mathbf{c} + \hat{\mathbf{F}}_{t\rightarrow t+\Delta}(\mathbf{c})\big)
\right),
\end{equation}
where $\hat{\mathbf{F}}_{t\rightarrow t+\Delta}$ denotes the predicted optical flow between two adjacent frames through an FPN-based flow decoder.
Here, $\mathbf{I}_t(\mathbf{c})$ and $\mathbf{I}_{t+\Delta}(\mathbf{c})$ represent the pixel intensities of two consecutive frames viewed as continuous functions of spatial coordinate $\mathbf{c}$.
The term $\mathbf{I}_{t+\Delta}(\mathbf{c} + \hat{\mathbf{F}}_{t\rightarrow t+\Delta}(\mathbf{c}))$ samples the next frame at a displaced location determined by the estimated flow, effectively warping it toward the current frame.
The robust function $\psi(\cdot)$ follows $(|u|+\epsilon)^q$ with $\epsilon=0.01$ and $q=0.4$~\cite{liu2019ddflow}.
For appearance modeling, a bounding box regression loss is given by
\begin{equation}
    \mathcal{L}_{\mathrm{bbox}} 
    = \| \hat{\mathbf{c}_\mathrm{b}} - \mathbf{c}_\mathrm{b} \|_1,
\end{equation}
where $\mathbf{c}_\mathrm{b}$ denotes the reference coordinates of bounding boxes.

Finally, we combine all losses, including the intra-modal contrastive loss $\mathcal{L}_{\mathrm{con}}$ (Sec.~\ref{sec:dual_disentangling}) and the cross-modal alignment losses $\mathcal{L}_{\mathrm{adv}}$ and $\mathcal{L}_{\mathrm{trans}}$ (Sec.~\ref{sec:cross_modal_alignment}), into the total objective as follows,
\begin{equation}
\begin{aligned}
    \mathcal{L}_{\mathrm{total}} &= 
    \mathcal{L}_{\mathrm{mov\_seg}}
    + \lambda_{\mathrm{flow}}\mathcal{L}_{\mathrm{flow}}
    + \lambda_{\mathrm{bbox}}\mathcal{L}_{\mathrm{bbox}} \\
    &\quad + \lambda_{\mathrm{con}}\mathcal{L}_{\mathrm{con}}
    + \lambda_{\mathrm{dist}}\mathcal{L}_{\mathrm{adv}} + \lambda_{\mathrm{sem}}\mathcal{L}_{\mathrm{trans}},
\end{aligned}
\end{equation}
where $\lambda_{\mathrm{flow}}$, $\lambda_{\mathrm{bbox}}$, $\lambda_{\mathrm{con}}$, $\lambda_{\mathrm{dist}}$, and $\lambda_{\mathrm{sem}}$ are balancing coefficients.

\section{Experiments}

In this section, we evaluate our proposed method (DIMOS) on three challenging datasets that contain both image and event modalities. We compare our approach with frame-based and event-assisted methods. We further provide ablation studies to analyze the contribution of each component in our architecture.

\begin{table}[htbp]
\centering
\caption{Summary of the three datasets}
\renewcommand{\arraystretch}{0.8} 
\begin{tabular}{lrrrr}
\toprule
Dataset & \begin{tabular}{@{}c@{}}Avg. Inst. \\ per Frame\end{tabular} & \begin{tabular}{@{}c@{}}Avg. Inst. \\ Mask Area\end{tabular} & \begin{tabular}{@{}c@{}}Avg. Foreg. \\ Mask Area\end{tabular} \\
\midrule
MouseSIS & $4.10$ & $0.73\%$ & $3.01\%$ \\
SEVD-Fixed & $7.68$ & $0.15\%$ & $1.12\%$ \\
EVIMO & $1.34$ & $3.74\%$ & $5.03\%$ \\
\bottomrule
\end{tabular}
\label{tab:dataset_stats}
\end{table}

\begin{table*}[t]
\centering
\caption{Quantitative comparison on MouseSIS, SEVD-Fixed, and EVIMO.}
\label{tab:quant_all}
\setlength{\tabcolsep}{8pt}
\renewcommand{\arraystretch}{0.9}
\begin{tabular}{l l l c c c c}
\toprule
Dataset & Method & Backbone & $\mathbf{mIoU}_{\mathrm{ins}}$ (\%) & $\mathbf{mIoU}_{01}$ (\%) & mAP (\%) & FLOPs \\
\midrule
\multirow{5}{*}{MouseSIS} 
& IDOL~\cite{wu2022defense}          & ResNet-50 & 60.66  & 66.96 & 26.73 & 68.13G \\
%& Segformer       & 54.47 & -- & -- & -- \\
& ModelMixSort~\cite{hamann2024mousesis}  & YOLO-SAM & \underline{63.72}  & \underline{75.79} & 23.11 & $5.49$T \\
& EvInsMOS~\cite{wan2025instance}      & ResNet-50 & 62.54  & 75.34 & \underline{30.94} & $26.08$G \\
\rowcolor{gray!10}
& DIMOS (ours)  & ResNet-50 & \textbf{70.25} & \textbf{77.30} & \textbf{45.18} & $60.42$G \\
\midrule
\multirow{5}{*}{SEVD-Fixed} 
& IDOL~\cite{wu2022defense}          & ResNet-50 & 45.49    & 52.17 & 16.13 & 195.85G \\
%& Segformer       & --    & -- & -- & -- \\
& ModelMixSort~\cite{hamann2024mousesis}  & YOLO-SAM & 49.56 & \underline{61.43} & 18.47 & $5.55$T \\
& EvInsMOS~\cite{wan2025instance}     & ResNet-50  & \underline{56.50} & 58.45 & \underline{20.24} & $87.52$G \\
\rowcolor{gray!10}
& DIMOS (ours) & ResNet-50 & \textbf{62.05} & \textbf{61.53} & \textbf{23.29} & $201.26$G \\
\midrule
\multirow{5}{*}{EVIMO} 
& IDOL~\cite{wu2022defense}         & ResNet-50  & 69.35 & 72.01 & 33.08 & 106.12G \\
%& Segformer       & 65.44 & 68.15 & -- & -- \\
& ModelMixSort~\cite{hamann2024mousesis}  & YOLO-SAM & \underline{71.67} & \textbf{78.33} & 33.99 & $5.50$T \\
& EvInsMOS~\cite{wan2025instance}     & ResNet-50  & 71.26 & 75.19 & \underline{35.97} & $40.95$G \\
\rowcolor{gray!10}
& DIMOS (ours) & ResNet-50 & \textbf{72.08} & \underline{75.74} & \textbf{36.44} & $94.81$G \\
\bottomrule
\end{tabular}
\end{table*}

\subsection{Datasets}

We conduct extensive experiments on three benchmarks: MouseSIS~\cite{hamann2024mousesis}, SEVD-Fixed~\cite{aliminati2024sevd}, and EVIMO~\cite{mitrokhin2019ev}.

\textbf{MouseSIS}~\cite{hamann2024mousesis} contains synchronized grayscale frames and event streams of interacting mice with over 75000 temporally consistent instance masks. The targets are small and frequently occluded, making it suitable for evaluating fine-grained segmentation of objects with low foreground ratio.

\textbf{SEVD-Fixed}~\cite{aliminati2024sevd} is a synthetic traffic surveillance dataset with RGB, event, depth, and semantic labels captured under diverse lighting and weather conditions. Foreground objects like vehicles and pedestrians are often small, challenging precise instance segmentation.

\textbf{EVIMO}~\cite{mitrokhin2019ev} provides indoor event streams with ground-truth motion masks and depth for up to three moving objects, serving as a standard benchmark for motion segmentation.

As shown in Table~\ref{tab:dataset_stats}, MouseSIS and SEVD-Fixed exhibit lower average instance mask area ratio (0.73\% and 0.15\%) compared to EVIMO (3.74\%), and lower foreground coverage. This highlights their particular challenge on small instance segmentation. More dataset details are provided in the supplementary material.

% \subsection{Comparison Methods}

% We compare our method with two categories of approaches:

% \textbf{Frame-based VIS.} We evaluate classical video instance segmentation methods such as IDOL~\cite{idol2022} and SegFormer~\cite{xie2021segformer}, which operate solely on image frames for high-quality segmentation under standard illumination.

% \textbf{Event-assisted fusion.} We also consider methods that combine event and image modalities, where event streams provide additional motion cues for improved temporal consistency and robustness in low-light conditions. Specifically, we adopt two representative event-assisted frameworks: \textit{ModelMixSort}, which leverages multimodal fusion to enhance frame-based segmentation by transforming event streams to gray images, and \textit{EvInsMOS}, which integrates multimodal representations by cross attention for robust moving instance segmentation.

\subsection{Implementation Details}

We implement our framework in PyTorch. For the three datasets, we train the network for 400K iterations on MouseSIS, 500K on EVIMO, and 800k on SEVD-Fixed with a batch size of 16. We use the Adam optimizer~\cite{kingma2014adam} with a weight decay of $1 \times 10^{-6}$ and employ a one-cycle learning rate schedule, with the peak learning rate set to $1 \times 10^{-4}$. The number of event bins is set to $B = 10$ and the moving confidence threshold is set to $\theta = 0.1$ across all experiments. The loss weights are set to $\lambda_{\mathrm{flow}} = 10.0$, $\lambda_{\mathrm{con}} = 0.5$, $\lambda_{\mathrm{bbox}} = 0.01$, $\lambda_{\mathrm{dist}} = 0.1$, and $\lambda_{\mathrm{sem}} = 10.0$. All experiments, including ablations and comparisons with prior methods, are conducted on the same evaluation machine. Training is performed on dual A40 GPUs, and inference is conducted on a single RTX 5090 GPU to ensure consistent evaluation settings. More architectural and training details are provided in the supplementary material. 
%All code is implemented in PyTorch and will be released upon acceptance.

Due to the relatively low spatial resolution of most existing DVS sensors and the requirement to maintain consistent resolution across modalities, we downsample different datasets to specific target resolutions. The input resolutions are resized to $320 \times 180$ for MouseSIS and $512 \times 384$ for SEVD-Fixed. This downsampling strategy significantly reduces computational overhead while preserving instance-level discriminability, particularly for small objects. For EVIMO, we use the original resolution of $346 \times 260$. Importantly, the lower resolution setting further highlights the challenge of small instance segmentation.

Following previous works~\cite{zhou2021event,lin2014microsoft,wan2025instance}, we adopt three primary metrics for moving instance segmentation: $\mathbf{mIoU}_{\mathrm{ins}}$, $\mathbf{mIoU}_{01}$, and $\mathbf{mAP}$. Specifically, $\mathbf{mIoU}_{\mathrm{ins}}$ evaluates instance-level segmentation accuracy for each moving object, while $\mathbf{mIoU}_{01}$ measures the 0–1 binary foreground mask accuracy. We further report $\mathbf{mAP}$ to account for false positives and overall detection precision.

\subsection{Quantitative Results}

We quantitatively evaluate the proposed framework against representative frame-based (IDOL~\cite{wu2022defense}) and event-assisted (ModelMixSort~\cite{hamann2024mousesis} and EvInsMos~\cite{wan2025instance}) methods on three challenging benchmarks: MouseSIS, SEVD-Fixed, and EVIMO. The results are summarized in Table~\ref{tab:quant_all}.

On the MouseSIS dataset, our method achieves the best instance-level segmentation accuracy with the $\mathbf{mIoU}_{\mathrm{ins}}$ of 70.25\%, outperforming both classical frame-based method (IDOL~\cite{wu2022defense}) and event-assisted baselines. Notably, ModelMixSort~\cite{hamann2024mousesis} and EvInsMOS~\cite{wan2025instance} already yield improvements over image-only methods, which confirms the benefits of leveraging event data under the challenging illumination condition. Our method further boosts performance through explicit disentanglement and alignment strategies. On the SEVD-Fixed dataset, which presents more challenging scenarios than MouseSIS due to its complex outdoor environments, diverse weather conditions, and a larger number of smaller instances, our framework achieves 62.05\% $\mathbf{mIoU}_{\mathrm{ins}}$, outperforming EvInsMOS~\cite{wan2025instance} by 5.55\%. This consistent gain highlights the superior robustness of our approach under extreme conditions, where event signals effectively complement degraded image data. On the EVIMO dataset, although EvInsMOS~\cite{wan2025instance} and ModelMixSort~\cite{hamann2024mousesis} already perform strongly by leveraging both modalities, our method achieves the highest $\mathbf{mIoU}_{\mathrm{ins}}$ of 72.08\%, indicating the effectiveness of our disentanglement strategy.

Overall, these results demonstrate that our method achieves consistent improvements across different benchmarks with diverse illumination conditions, motion patterns, and scene complexities. Importantly, both MouseSIS and SEVD-Fixed contain a large number of small instances, where accurate segmentation strongly relies on dense appearance and motion features. Our disentanglement framework effectively enhances the segmentation of such small instances by simultaneously extracting appearance and motion information from both modalities. In contrast, conventional frame-based or simple fusion methods do not explicitly perform such dual-modality disentangling, which highlights the advantage of our approach in jointly leveraging complementary appearance and motion cues.

\subsection{Qualitative Results}
% (Section intentionally left blank for now)
% TODO: include visualization of small object segmentation under low light

\begin{figure*}[t]
  \centering
   \includegraphics[width=1.0\linewidth]{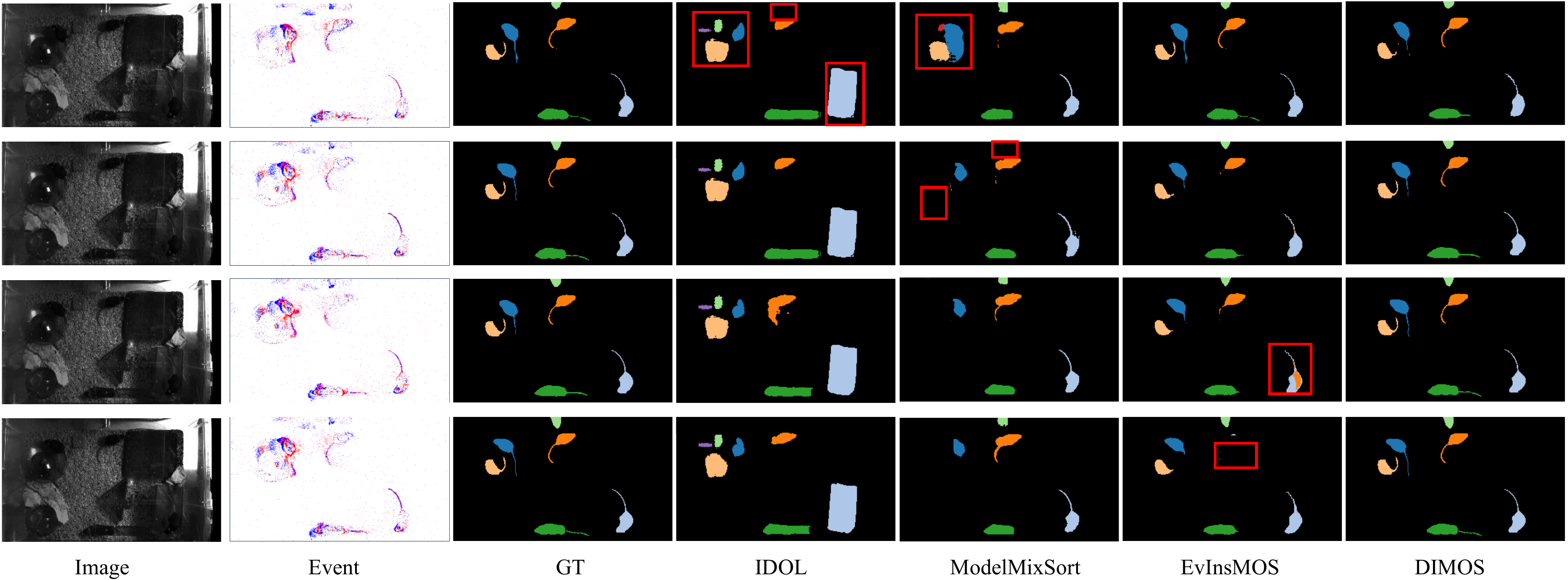}

\caption{Visual comparisons show consecutive frames sampled from a video sequence of MouseSIS~\cite{hamann2024mousesis}, arranged from top to bottom. Red boxes highlight regions with missed segmentation or unclear boundaries.
}
\label{fig:qualitative}
\end{figure*}
Figure~\ref{fig:qualitative} presents qualitative comparisons among representative methods on MouseSIS datasets. The results consistently demonstrate that our proposed DIMOS framework achieves more accurate and temporally consistent segmentation, particularly for small moving instances.

Compared to IDOL, our model produces cleaner object boundaries and avoids missing detections under motion blur or low-illumination conditions. The superiority becomes evident where image-only models often fail to distinguish object contours or confuse overlapping instances. Event-assisted baselines, including ModelMixSort~\cite{hamann2024mousesis} and EvInsMOS~\cite{wan2025instance}, perform better by leveraging event information; however, they still exhibit fragmented masks or inaccurate separations when multiple objects move closely or interact. Overall, the visual comparisons align well with the quantitative results, confirming that the proposed framework achieves superior robustness for small instance segmentation under limited resolution. 
%DIMOS not only captures detailed object boundaries but also maintains instance integrity and stability across time, highlighting its effectiveness for challenging moving instance segmentation tasks in both indoor and outdoor scenes.

\subsection{Ablation Study}

% \begin{table}[ht]
% \centering
% \caption{Ablation study on MouseSIS. We incrementally add dual-disentangling mechanism (dual. mech.), extra task-specific loss (UnFlow and BBox), feature alignment (Feat. Align.), and distribution alignment (Dist. Align.) on top of different extra-task settings.}
% \label{tab:ablation}
% \setlength{\tabcolsep}{6pt}
% \begin{tabular}{ccccccc}
% \toprule
% UnFlow & BBox & Dual. mech. & Feat. Align. & Dist. Align. & mIoU (\%) \\
% \midrule
% \(\times\) & \(\times\) & \(\times\) & \(\times\) & \(\times\) &  58.4 \\
% \(\checkmark\) & \(\times\) & \(\times\) & \(\times\) & \(\times\) & 60.8 \\
% \(\checkmark\) & \(\checkmark\) & \(\times\) & \(\times\) & \(\times\) & 62.4 \\
% \(\checkmark\) & \(\checkmark\) & \(\checkmark\) & \(\times\) & \(\times\) & 68.1 \\
% \(\checkmark\) & \(\checkmark\) & \(\checkmark\) & \(\checkmark\) & \(\times\) & 69.2 \\
% \(\checkmark\) & \(\checkmark\) & \(\checkmark\) & \(\checkmark\) & \(\checkmark\) & 70.2 \\
% \bottomrule
% \end{tabular}
% \end{table}

\begin{table}[ht]
\centering
\caption{Ablation study on MouseSIS. We incrementally add dual-disentangling mechanism (Dual. Mech.), extra task-specific losses (UnFlow and BBox), semantic alignment loss (Sem. Align.), and distributional alignment loss (Dist. Align.).}
\label{tab:ablation}
\setlength{\tabcolsep}{5pt} % 减少列间距
\begin{tabular}{ccccccc}
\toprule
\makecell{UnFlow\\} & 
\makecell{BBox\\} & 
\makecell{Dual.\\Mech.} & 
\makecell{Sem.\\Align.} & 
\makecell{Dist.\\Align.} & 
\makecell{$\mathbf{mIoU}_{\mathrm{ins}}$\\(\%)} \\
\midrule
\(\times\) & \(\times\) & \(\times\) & \(\times\) & \(\times\) &  60.47 \\
\(\checkmark\) & \(\times\) & \(\times\) & \(\times\) & \(\times\) & 62.54 \\
\(\checkmark\) & \(\checkmark\) & \(\times\) & \(\times\) & \(\times\) & 63.46 \\
\(\checkmark\) & \(\checkmark\) & \(\checkmark\) & \(\times\) & \(\times\) & 68.11 \\
\(\checkmark\) & \(\checkmark\) & \(\checkmark\) & \(\checkmark\) & \(\times\) & 69.23 \\
\(\checkmark\) & \(\checkmark\) & \(\checkmark\) & \(\checkmark\) & \(\checkmark\) & 70.25 \\
\bottomrule
\end{tabular}
\end{table}

To investigate the contribution of each component, we perform ablation experiments on MouseSIS by progressively enabling task-specific supervision, dual-disentangling mechanism, alignment modules, and encoder backbones. Results are shown in Table~\ref{tab:ablation} and Table~\ref{tab:backbone}.

\textbf{Baseline without extra modules.}
As shown in Table~\ref{tab:ablation}, without any additional modules, the model reduces to a simple multimodal interaction pipeline, achieving only 60.47\% $\mathbf{mIoU}_{\mathrm{ins}}$, which confirms the weakness of insufficient feature extraction in handling small instance segmentation under challenging conditions.

\textbf{Effect of extra task-specific supervision.} 
As shown in Table~\ref{tab:ablation}, introducing unsupervised flow estimation improves $\mathbf{mIoU}_{\mathrm{ins}}$ from 60.47\% to 62.54\%, demonstrating that additional motion guidance helps capture motion cues. Incorporating bounding box supervision further raises it to 63.46\%, providing a spatial prior that enhances localization, particularly for small or overlapping objects. For sequences without box annotations, pseudo boxes are generated from the outer boundaries of instance masks.

\textbf{Effect of dual-disentangling mechanism.}
Table~\ref{tab:ablation} also shows that the dual-disentangling mechanism brings a significant performance increase to 68.11\% $\mathbf{mIoU}_{\mathrm{ins}}$. This module explicitly separates appearance and motion information within both modalities, resulting in denser and more discriminative features after cross-modal fusion, which are particularly beneficial for small instance segmentation. By introducing intra-modal contrastive learning for explicit disentanglement, the network prevents the mutual interference between appearance and motion semantics and ensures that each branch focuses on its corresponding representation.

\textbf{Effect of semantic and distributional alignment.}
Adding the semantic-level alignment further improves the performance to 69.23\% $\mathbf{mIoU}_{\mathrm{\mathrm{ins}}}$, indicating that transforming corresponding appearance and motion features across modalities enhances fusion effectiveness. Moreover, enabling distribution alignment pushes the performance to 70.25\% $\mathbf{mIoU}_{\mathrm{ins}}$ since features extracted from image and event streams naturally follow different distributions that may degrade performance. The semantic reconstruction and adversarial domain adaptation jointly ensure that representations from different modalities remain coherent.

\begin{table}[ht]
\centering
\caption{Backbone ablation on MouseSIS dataset.}
\label{tab:backbone}
\setlength{\tabcolsep}{4pt}
\begin{tabular}{lccc}
\toprule
Backbone & Param. & FLOPs & $\mathbf{mIoU}_{\mathrm{ins}}$ (\%)  \\
\midrule
MobileNetV2~\cite{sandler2018mobilenetv2} & $\sim3.4$M & $12.24$G & 68.62  \\
ResNet18~\cite{he2016deep} & $\sim11.7$M & $20.10$G & 69.32  \\
ResNet-50~\cite{he2016deep} & $\sim25.6$M & $60.42$G & \textbf{70.25} \\
\bottomrule
\end{tabular}
\end{table}

\textbf{Effect of different encoder backbone.}  
Table~\ref{tab:backbone} shows the results using different backbones on MouseSIS, confirming that our disentangling and alignment modules effectively leverage the representational capacity of deeper networks. Importantly, using lightweight backbones such as MobileNetV2~\cite{sandler2018mobilenetv2} and ResNet-18~\cite{he2016deep} results in only marginal drops of 1.63\% and 0.93\%, demonstrating the strong backbone-agnostic generalization of our framework. This shows that performance gains come from the proposed modules rather than large encoders alone. This is particularly advantageous given that our disentangling framework involves multiple encoder branches. By adopting dual lightweight backbone networks, we can reduce overall parameters while maintaining or even surpassing the performance of conventional methods that rely on a single, large-capacity backbone (e.g., dual MobileNetV2 with around 7.0M parameters vs. single ResNet-50~\cite{he2016deep} with around 25.6M parameters), achieving a favorable performance–efficiency trade-off.

\section{Conclusion}
In this work, we addressed the challenge of small moving instance segmentation by proposing the DIMOS framework. Our method disentangles and extracts appearance and motion representations within each modality and aligns them through a multi-granularity cross-modal alignment strategy. This design enhances feature density and fusion effectiveness, leading to consistent improvements across multiple datasets. Experimental results demonstrate the effectiveness and robustness of our approach, particularly for small instances under challenging conditions. 

\noindent\textbf{Limitation.} Our framework, like most multimodal segmentation systems, still relies on paired inputs. However, such synchronized dual-modality inputs are not always available. Existing methods often experience severe performance degradation or even fail completely in such single-modality settings. Therefore, enhancing the single-modality compatibility of multimodal systems represents an important and meaningful future research direction.

%However, the improvement on objects with large foreground regions remains relatively limited. This is mainly because such regions inherently contain richer appearance and motion information, which reduces the relative advantage of our disentangling and alignment strategies. Moreover, the simplicity of the current feature representations prevents the model from fully exploiting these abundant cues, limiting its ability to model fine-grained motion and structural variations. In the future, we plan to explicitly incorporate multiple representations per modality for exploring richer appearance and motion cues, to enhance the generalization capability of our framework.
\section{Acknowledgements}
This work was partially supported by the Young Scientists Fund of the National Natural Science Foundation of China under Grant (62305278), as well as the Youth S\&T Talent Support Program of GDSTA under Grant (SKXRC2025460). The authors gratefully acknowledge the financial support that made this research possible.
\clearpage
\setcounter{page}{1}
\maketitlesupplementary
\setcounter{section}{0}
\renewcommand{\thesection}{\Alph{section}}
\renewcommand{\thesubsection}{\thesection.\arabic{subsection}}

\begin{figure}[t]
  \centering
   \includegraphics[width=1.0\linewidth]{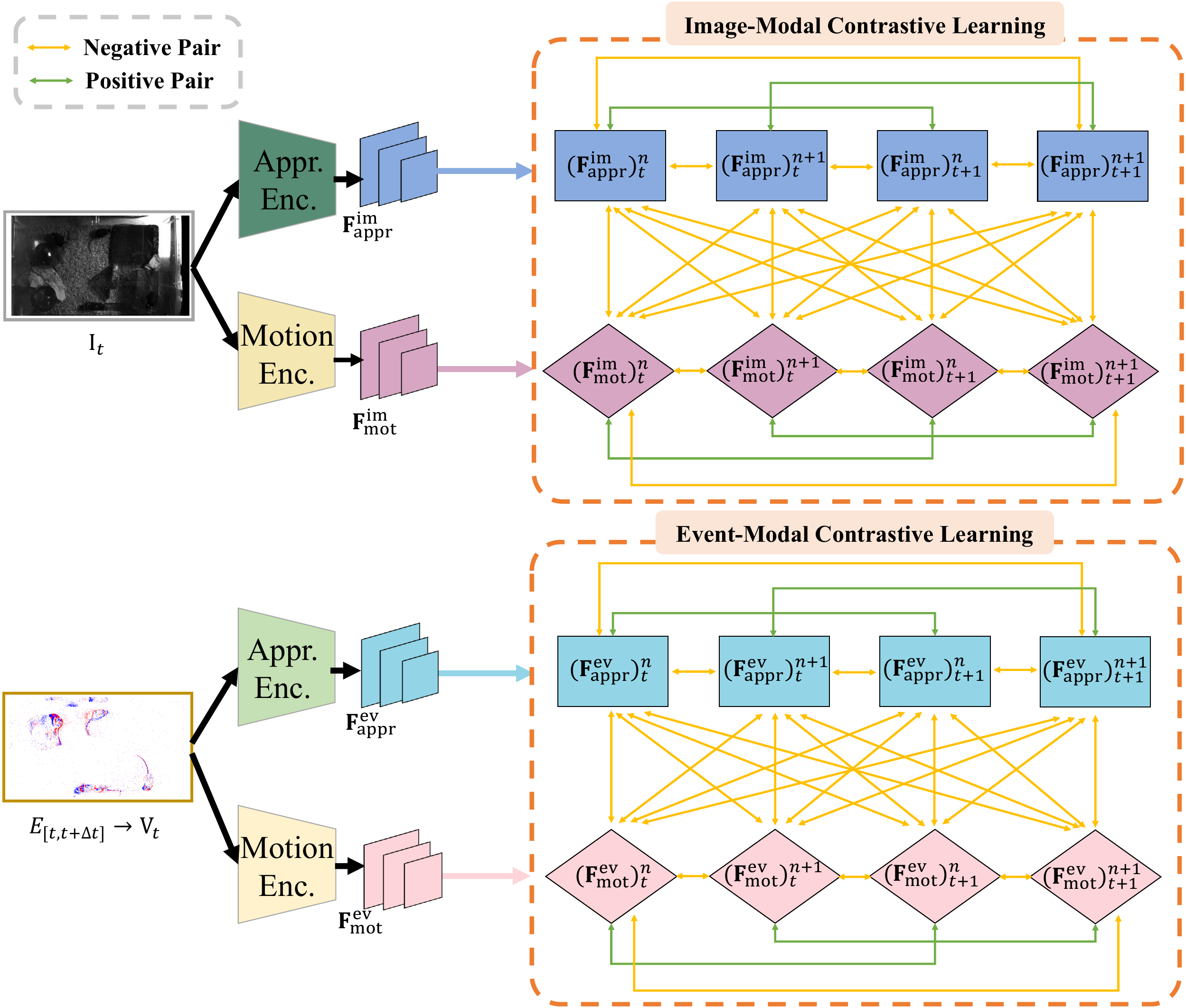}
  \caption{
Intra-modal contrastive learning for strengthening appearance--motion disentanglement.
}
\label{fig:contrast_construction}
\end{figure}

\section{Detailed Network Architecture}
To provide a more comprehensive understanding of our multimodal fusion framework, we complement the descriptions in the main paper by elaborating the architectural details of the modules introduced for disentangled cross-modal fusion and multi-granularity alignment. As described in Sec.~\ref{sec:dual_disentangling} of the main paper, the dual-disentangling mechanism forms the backbone of our feature extraction pipeline, where each modality is equipped with a dual-branch encoder that separates appearance-related and motion-related representations. Building upon these disentangled features, we incorporate a lightweight fusion module operating across multiple spatial scales, along with a series of auxiliary alignment modules that act at different granularity levels. %These components are instrumental during training but are designed to be discarded during inference, ensuring that the proposed framework retains a compact and efficient test-time architecture.

\begin{table*}[t]
\centering
\caption{Network structure of the Cross-Modal Fusion Module used for merging same-type features 
from image and event modalities. Three parallel fusion layers operate on three 
feature pyramid levels independently.}
\begin{tabular}{c|c|c}
\hline
\textbf{Scale Level} & \textbf{Input} & \textbf{Cross-Modal Fusion Block (per level)} \\ \hline
Level 1 
& $\mathrm{Concat}(\mathbf{F}^{a}_1, \mathbf{F}^{b}_1)$ 
& Conv2d($2C_1$, $C_1$, $3\times3$, pad=1) + ReLU \\ \hline

Level 2 
& $\mathrm{Concat}(\mathbf{F}^{a}_2, \mathbf{F}^{b}_2)$ 
& Conv2d($2C_2$, $C_2$, $3\times3$, pad=1) + ReLU \\ \hline

Level 3 
& $\mathrm{Concat}(\mathbf{F}^{a}_3, \mathbf{F}^{b}_3)$ 
& Conv2d($2C_3$, $C_3$, $3\times3$, pad=1) + ReLU \\ \hline
\end{tabular}
\end{table*}

\begin{table*}[t]
\centering
\caption{Network architectures of the extra modules used for multi-granularity 
cross-modal alignment, including domain discriminators for distribution-level 
alignment and translators for semantic-level alignment. All modules are used only 
during training.}
\begin{tabular}{c|c|c}
\hline
\textbf{Module} & \textbf{Input} & \textbf{Network Block} \\ \hline

Domain Discriminator 
& Feature map $\mathbf{F}$ 
& \begin{tabular}{@{}c@{}}
Conv2d($C$, 64, $3\times3$, pad=1) \\
BatchNorm2d(64) + ReLU \\
Conv2d(64, 1, $1\times1$)
\end{tabular} 
\\ \hline

Modality Translator 
& Event feature $\mathbf{F}^{\mathrm{ev}}$ 
& \begin{tabular}{@{}c@{}}
Conv2d($C$, $C$, $3\times3$, pad=1) + ReLU \\
Conv2d($C$, $C'$, $1\times1$)
\end{tabular} 
\\ \hline

\end{tabular}
\end{table*}

\subsection{Fusion Module for Disentangled Cross-modal Features}
The dual-branch encoders applied to image frames and event streams respectively produce appearance features 
$\mathbf{F}^{\mathrm{im}}_{\mathrm{appr}}$ and $\mathbf{F}^{\mathrm{ev}}_{\mathrm{appr}}$, as well as motion features 
$\mathbf{F}^{\mathrm{im}}_{\mathrm{mot}}$ and $\mathbf{F}^{\mathrm{ev}}_{\mathrm{mot}}$. 
Each branch generates a pyramid of feature maps at three different resolutions, reflecting the multi-scale hierarchy established by the encoder design.  
The goal of the fusion module is to merge same-type features originating from different modalities. For this purpose, we adopt a compact architecture built from three parallel convolutional layers, each responsible for one scale of the pyramid. 

At every scale, the fusion process is performed by concatenating features from the two modalities, followed by a $3 \times 3$ convolution with a ReLU nonlinearity:
\[
    \mathbf{F}^{\mathrm{fused}} = 
    \phi\big(\mathrm{Conv}_{3 \times 3}(
    \mathrm{Concat}(\mathbf{F}_a, \mathbf{F}_b))\big),
\]
where $\phi(\cdot)$ denotes the ReLU activation. 
This simple yet effective design allows the fused features to integrate complementary information from image and event inputs, while simultaneously maintaining computational efficiency. Because the encoder produces features at three hierarchical levels, the fusion module mirrors this structure by employing three independent convolutional units, ensuring that each resolution is processed in a scale-aware manner. This design choice enables the downstream decoder to operate on features that are both semantically coherent across modalities and structurally aligned across scales.

\subsection{Extra Modules for Multi-Granularity Cross-Modal Alignment}
Beyond the cross-modal fusion, our framework benefits from additional modules that impose alignment constraints between image and event representations. These modules operate at two distinct granularity levels and are used exclusively during the training phase. Their primary purpose is to promote consistent distributions and semantics across modalities, thereby enhancing the efficacy of the fused features introduced above.

At the distribution level, the model incorporates discriminators that implement an adversarial learning scheme to reduce the discrepancy between feature distributions of the two modalities. For each feature type, whether appearance or motion, we employ an independent pixel-wise discriminator equipped with a single convolutional layer followed by BatchNorm, ReLU, and a final $1 \times 1$ convolution that outputs spatially resolved domain confidence scores. Despite being lightweight, these discriminators effectively capture modality-specific biases and encourage the encoders to produce domain-invariant representations under the adversarial objective described in Sec.~\ref{sec:cross_modal_alignment} of the main paper.

At the semantic level, we further introduce lightweight feature translators that promote fine-grained alignment by explicitly transforming event features toward the image domain (or vice versa). Each translator consists of two convolutional layers: a $3 \times 3$ layer refining local structure, and a $1 \times 1$ layer adjusting the channel configuration to match the target representation. These translators provide an additional supervisory signal that complements the adversarial alignment, helping ensure that the fused features preserve modality-consistent semantics even in the presence of appearance–motion disentanglement. As with the discriminators, the translators are required only during training and impose no computational burden at inference time.

\section{Construction of Positive and Negative Samples for Intra-modal Contrastive Learning}
As shown in Figure~\ref{fig:contrast_construction}, our contrastive learning formulation operates entirely within the same modality rather than across modalities. For a batch of samples indexed by $n$, each sample consists of two consecutive frames (or event windows) at times $t$ and $t{+}1$, from which the dual-branch encoder extracts disentangled appearance features $(\mathbf{F}_{\mathrm{appr}})^n_t$, $(\mathbf{F}_{\mathrm{appr}})^n_{t+1}$ and motion features $(\mathbf{F}_{\mathrm{mot}})^n_t$, $(\mathbf{F}_{\mathrm{mot}})^n_{t+1}$. Positive pairs are formed by associating features of the \emph{same branch type} (appearance--appearance or motion--motion) across the two consecutive frames of the same sample, capturing temporal continuity within each semantic space. Negative pairs are obtained either from features of \emph{different branch types} within the same sample (appearance vs.\ motion) or from temporally non-adjacent samples across the batch. This intra-modal formulation compels the network to separate appearance and motion cues within each modality, preventing representational mixing across the two branches and producing cleaner disentangled feature spaces.

\begin{figure}[t]
  \centering
   \includegraphics[width=1.0\linewidth]{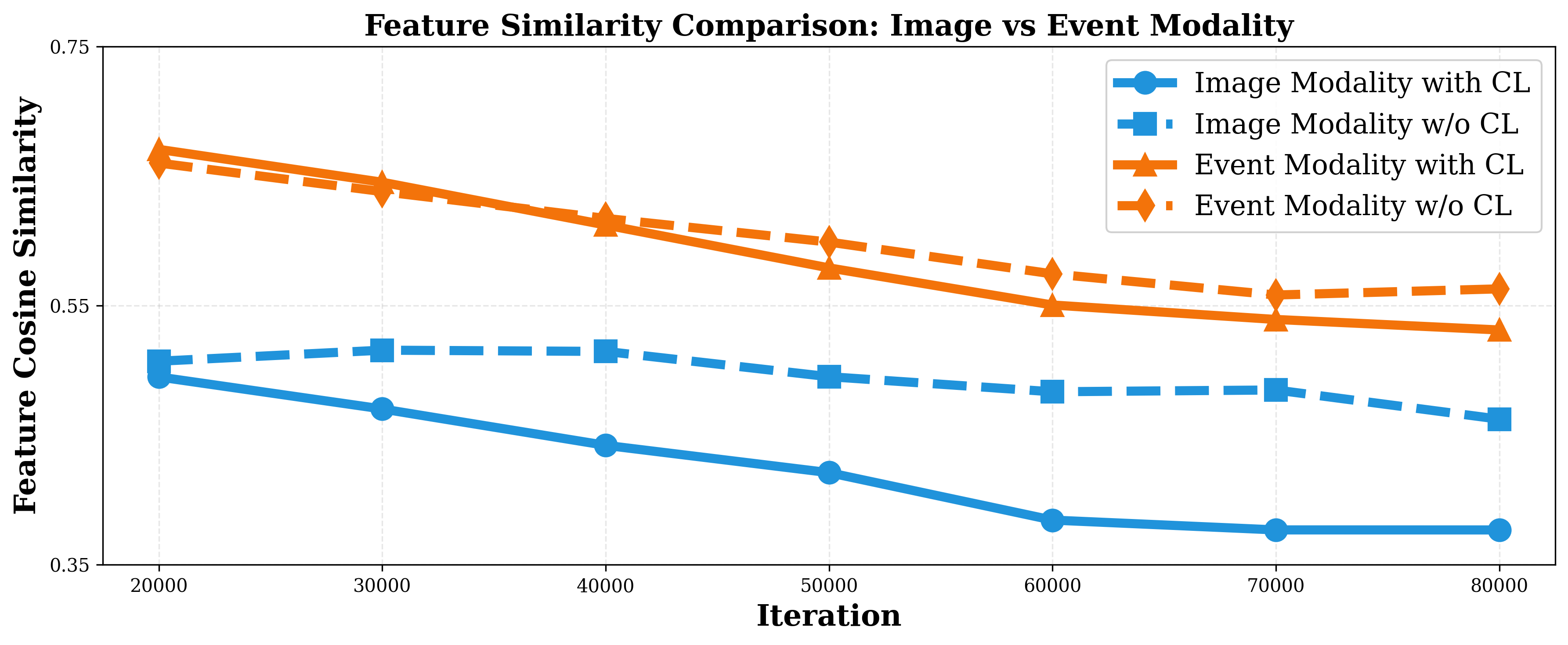}
  \caption{
Effect of intra-modal contrastive learning on appearance-and-motion feature similarity of each modality.
}
\label{fig:intra_modal_similarity}
\end{figure}

We further quantify the effect of intra-modal contrastive learning by measuring the cosine similarity between appearance and motion features during training. As shown in Figure~\ref{fig:intra_modal_similarity}, applying contrastive learning leads to a clear downward trend in similarity for both image and event modalities, demonstrating that the appearance and motion branches become increasingly decorrelated as training progresses. These results provide direct empirical evidence that the proposed intra-modal contrastive objective effectively strengthens appearance-and-motion disentanglement within each modality.

\section{Detailed Formula for Distribution-level Alignment}
For completeness, we present the full formulation of the adversarial loss used to perform distribution-level alignment, following the notation in Sec.~\ref{sec:cross_modal_alignment} of the main paper.  

For appearance features extracted from image and event modalities, 
$\mathbf{F}^{\mathrm{im}}_{\mathrm{appr}}$ and $\mathbf{F}^{\mathrm{ev}}_{\mathrm{appr}}$, 
the adversarial loss associated with the appearance discriminator $D_a$ is expressed as
\[
\begin{aligned}
\mathcal{L}^{\mathrm{appr}}_{\mathrm{adv}}
=\;&
\mathbb{E}_{\mathbf{F}^{\mathrm{im}}_{\mathrm{appr}}}
\!\left[
    \log D_a(\mathbf{F}^{\mathrm{im}}_{\mathrm{appr}})
\right]
\\
&+
\mathbb{E}_{\mathbf{F}^{\mathrm{ev}}_{\mathrm{appr}}}
\!\left[
    \log \!\left( 1 - D_a(\mathbf{F}^{\mathrm{ev}}_{\mathrm{appr}}) \right)
\right].
\end{aligned}
\]
The discriminator attempts to maximize this quantity by differentiating between image and event appearance features, while the encoders minimize it through gradient reversal, thereby encouraging modality-invariant appearance representations.

A parallel formulation governs the motion features 
$\mathbf{F}^{\mathrm{im}}_{\mathrm{mot}}$ and $\mathbf{F}^{\mathrm{ev}}_{\mathrm{mot}}$, 
with motion discriminator $D_m$:
\[
\begin{aligned}
\mathcal{L}^{\mathrm{motion}}_{\mathrm{adv}}
=\;&
\mathbb{E}_{\mathbf{F}^{\mathrm{ev}}_{\mathrm{mot}}}
\!\left[
    \log D_m(\mathbf{F}^{\mathrm{ev}}_{\mathrm{mot}})
\right]
\\
&+
\mathbb{E}_{\mathbf{F}^{\mathrm{im}}_{\mathrm{mot}}}
\!\left[
    \log \!\left( 1 - D_m(\mathbf{F}^{\mathrm{im}}_{\mathrm{mot}}) \right)
\right].
\end{aligned}
\]
The combined adversarial objective used during training is then given by
\[
    \mathcal{L}_{\mathrm{adv}}
    =
    \mathcal{L}^{\mathrm{appr}}_{\mathrm{adv}}
    +
    \mathcal{L}^{\mathrm{mot}}_{\mathrm{adv}}.
\]
This loss encourages the encoders to align feature distributions across the two modalities at both appearance and motion levels.

\begin{figure*}[t]
  \centering
   \includegraphics[width=1.0\linewidth]{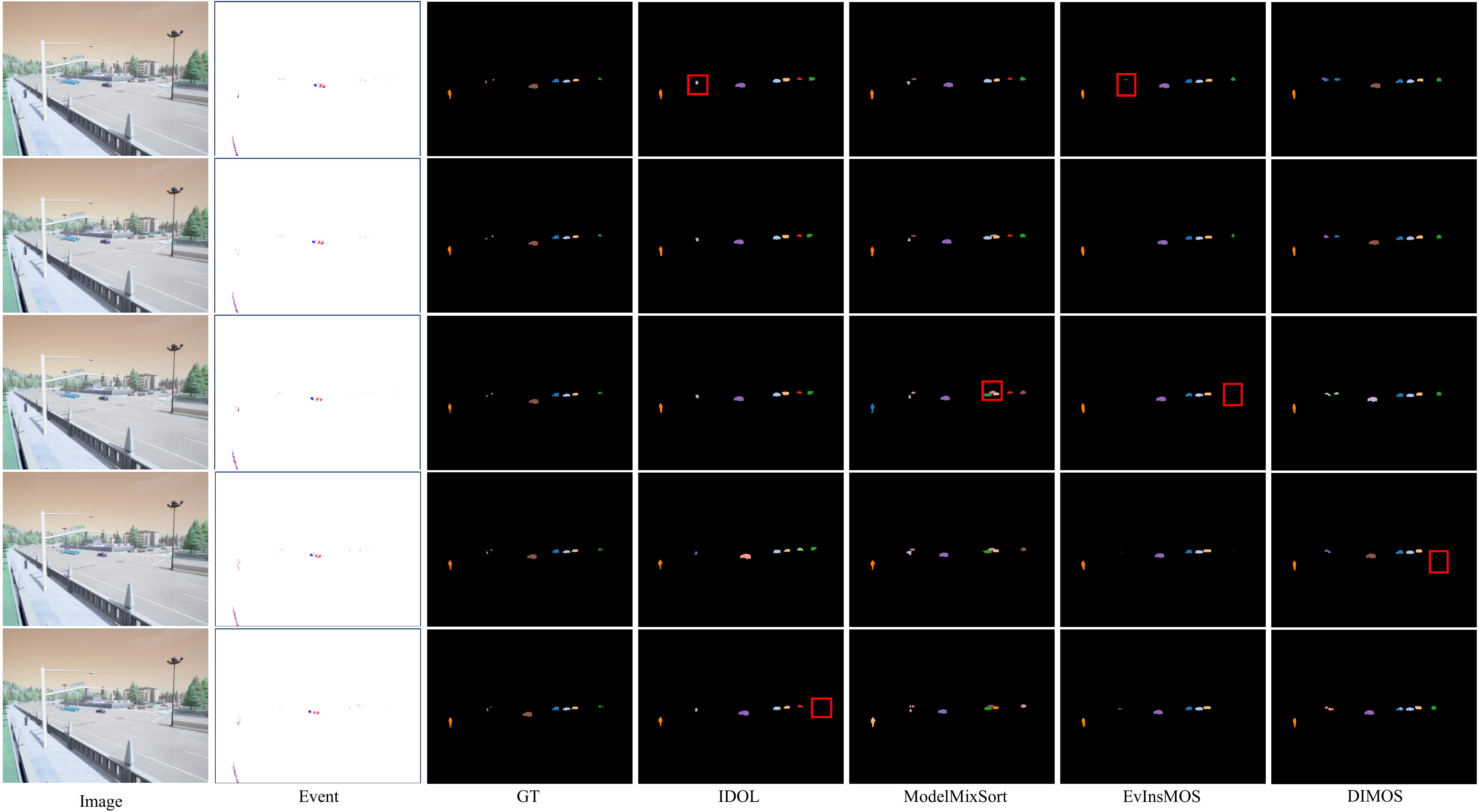}

\caption{Consecutive frames sampled from a video sequence of SEVD-Fixed are arranged from top to bottom. Red boxes highlight regions where competing methods miss small objects or fail to delineate clear boundaries.
}
\label{fig:sevd_vis}
\end{figure*}

\section{Dataset Details and Preprocessing}

To provide a clearer understanding of the data used throughout our experiments, we elaborate on the characteristics of the three multimodal datasets adopted in the main paper and describe the preprocessing procedures required to prepare the image--event pairs for our disentangled multimodal segmentation framework. Compared with the condensed description in the main paper, this section offers a more comprehensive account of the sensing configurations, annotation formats, temporal synchronization, and data transformations applied prior to training.

\paragraph{MouseSIS.}
MouseSIS is a recently introduced dataset designed for space--time instance segmentation of interacting laboratory mice. It offers paired \emph{grayscale image frames} and \emph{event streams} captured simultaneously through a beam-splitter optical assembly, which ensures that both modalities share an almost perfectly aligned field of view. The recording system provides high spatial resolution for the grayscale camera and preserves the microsecond-level temporal precision inherent to the event sensor, allowing the dataset to capture fast body movements, occlusions, and fine-scale interactions among multiple mice. The dataset contains approximately $75{,}000$ instance masks with consistent identity annotations across time, enabling high-quality evaluation of both appearance- and motion-dependent segmentation cues.  
Two illumination conditions (1100 and 660 lux) are included, which introduce substantial visual variability in texture sharpness and event firing patterns. The event streams exhibit strong motion cues but relatively sparse spatial support, particularly under low-motion or low-contrast scenarios. To ensure consistent processing, events are first synchronized to the image timestamps using their inherent microsecond stamps, and subsequently accumulated into voxel grids over short temporal windows. The grayscale frames undergo standard intensity normalization, and both modalities are resized to a unified spatial resolution compatible with the dual-branch encoders.

\paragraph{SEVD-Fixed.}
SEVD-Fixed is derived from the SEVD dataset generated using the CARLA simulator and constitutes a diverse synthetic benchmark for multimodal understanding in traffic environments. The fixed-perception setting provides four static camera viewpoints deployed at intersections and roundabouts, each of which simultaneously produces \emph{RGB frames}, \emph{event streams}, \emph{depth maps}, \emph{semantic labels}, and \emph{instance-level masks}.  
The scenes cover a broad spectrum of variations in both lighting (noon, night, twilight) and weather conditions (clear, cloudy, rainy, foggy), generating substantial appearance diversity that is challenging for many event-based models. The event streams are produced directly within the CARLA rendering engine by simulating brightness-change triggers, allowing deterministic alignment between image frames and event packets.

Although SEVD-Fixed provides high-quality instance-level masks, these annotations describe \emph{all} objects in the scene, including static structures such as buildings and poles. Because the dataset itself does not contain explicit ground-truth masks for \emph{moving} objects, an additional procedure is required to isolate dynamic instances for our motion segmentation task. To this end, we make use of the motion-object detection boxes provided in the dataset and treat them as the reference modality for identifying dynamic entities. For each image frame, the detection boxes are matched with the full-scene instance segmentation masks using an IoU-based Hungarian assignment~\cite{cheng2021per}, ensuring a one-to-one and globally optimal correspondence between detections and instance masks.  
Through this matching process, we extract the subset of instance masks corresponding to actual moving objects, thereby constructing accurate motion-instance annotations from the originally static-inclusive segmentation labels. This conversion yields clean supervision tailored to the motion-centric objectives of our framework while maintaining full compatibility with the original dataset structure.

For preprocessing, RGB images are converted to a normalized imaging modality consistent with the other datasets, while events are aggregated into voxel tensors using a fixed number of temporal bins.

\paragraph{EVIMO.}
EVIMO is an indoor dataset specifically constructed for evaluating motion segmentation under independent object motion. It provides \emph{grayscale image frames}, \emph{event streams}, \emph{depth}, and \emph{pixel-level ground-truth motion masks}. Recordings span approximately $32$ minutes and capture up to three independently moving objects per sequence. 
The ground-truth motion masks are computed using a dedicated motion-capture pipeline and thereby provide highly accurate supervision for evaluating the dynamic branch of our network. To align the modalities, event packets are integrated over short windows centered around each image timestamp, producing voxel representations that preserve both the temporal density and spatial sparsity characteristic of event measurements.

\section{Extended Ablation Studies}
To further validate the effectiveness of the components introduced in our framework, we present a series of extended ablation experiments focusing specifically on the newly proposed modules. 
%and loss terms. These analyses isolate the contribution of each design choice, such as cross-modal fusion depth, distribution- and semantic-level alignment, and auxiliary regularization, to the overall performance. For all other hyperparameters and training settings, we follow the standard configurations established in prior multimodal event–image segmentation works~\cite{wan2025instance,liu2019ddflow}, ensuring that the observed improvements stem solely from the innovations introduced in this study.

\subsection{Depth of the Cross-Modal Fusion Module}
Table~\ref{tab:fusion_blocks} investigates how the number of blocks in the Cross-Modal Fusion Module affects the final performance. Using a single fusion block already brings strong performance, while stacking additional blocks yields marginal gains and even slight degradation in both $\mathrm{mIoU}_{\mathrm{ins}}$ and $\mathrm{mIoU}_{01}$. This result suggests that a shallow fusion design is sufficient to integrate complementary cues from images and events.

\subsection{Depth of the Domain Discriminator}
Table~\ref{tab:disc_blocks} reports the influence of the discriminator depth for distribution-level alignment. A single-block discriminator achieves the best results, while deeper variants consistently reduce both metrics. This indicates that an overly strong discriminator may overfit domain-specific patterns and hinder stable adversarial training, whereas a lightweight pixel-wise discriminator is sufficient to enforce the desired modality-invariant distributions.

\subsection{Depth of the Modality Translator}
Table~\ref{tab:trans_blocks} examines the number of blocks in the Modality Translator used for semantic-level alignment. All configurations with one to three blocks yield very similar performance. This demonstrates that our translation-based alignment is relatively insensitive to the exact depth, and that the single-block design used in the main model strikes a good balance between accuracy and efficiency.

\begin{table}[t]
\centering
\caption{Ablation study on the number of blocks used in the 
\textbf{Fusion Module for Disentangled Cross-modal Features}.}
\label{tab:fusion_blocks}
\begin{tabular}{lccc}
\toprule
\textbf{Network Module} & 
\textbf{\# Blocks} & 
\textbf{mIoU\textsubscript{ins}} & 
\textbf{mIoU\textsubscript{01}} \\
\midrule
Cross-Modal Fusion Module & 1 & 70.25 & 77.30 \\
Cross-Modal Fusion Module & 2 & 70.15 & 77.08 \\
Cross-Modal Fusion Module & 3 & 70.23 & 77.27 \\
\bottomrule
\end{tabular}
\end{table}

\begin{table}[t]
\centering
\caption{Ablation study on the number of blocks used in the 
\textbf{Domain Discriminator} for distribution-level alignment.}
\label{tab:disc_blocks}
\begin{tabular}{lccc}
\toprule
\textbf{Network Module} & 
\textbf{\# Blocks} & 
\textbf{mIoU\textsubscript{ins}} & 
\textbf{mIoU\textsubscript{01}} \\
\midrule
Domain Discriminator & 1 & 70.25 & 77.30 \\
Domain Discriminator & 2 & 69.89 & 76.97 \\
Domain Discriminator & 3 & 69.52 & 76.71 \\
\bottomrule
\end{tabular}
\end{table}

\begin{table}[t]
\centering
\caption{Ablation study on the number of blocks used in the 
\textbf{Modality Translator} for semantic-level alignment.}
\label{tab:trans_blocks}
\begin{tabular}{lccc}
\toprule
\textbf{Network Module} & 
\textbf{\# Blocks} & 
\textbf{mIoU\textsubscript{ins}} & 
\textbf{mIoU\textsubscript{01}} \\
\midrule
Modality Translator & 1 & 70.25 & 77.30 \\
Modality Translator & 2 & 70.27 & 77.11 \\
Modality Translator & 3 & 70.11 & 77.21 \\
\bottomrule
\end{tabular}
\end{table}

\section{Extended Quantitative Results}

To further complement the quantitative and qualitative analysis presented in the main paper, we provide additional visualization results on the \textbf{SEVD-Fixed} dataset. Compared with MouseSIS, SEVD-Fixed contains a significantly larger number of instances per frame and a notably higher proportion of small and distant objects. These characteristics make the segmentation task inherently more challenging, often leading to missed detections, incomplete boundaries, or fragmented masks across all compared approaches.

As shown in Figure~\ref{fig:sevd_vis}, the increased object density and reduced object size amplify the difficulty for existing multimodal and event-based segmentation frameworks. Our proposed DIMOS framework demonstrates stronger robustness against under-segmentation, yielding more complete and stable predictions even when the visual and event cues become sparse or noisy. The improvements are especially evident in the highlighted regions, where competing methods frequently miss small moving objects or merge them with the background, while DIMOS maintains clearer instance separation.

% \section{Rationale}
% \label{sec:rationale}
% % 
% Having the supplementary compiled together with the main paper means that:
% % 
% \begin{itemize}
% \item The supplementary can back-reference sections of the main paper, for example, we can refer to \cref{sec:intro};
% \item The main paper can forward reference sub-sections within the supplementary explicitly (e.g. referring to a particular experiment); 
% \item When submitted to arXiv, the supplementary will already included at the end of the paper.
% \end{itemize}
% % 
% To split the supplementary pages from the main paper, you can use \href{https://support.apple.com/en-ca/guide/preview/prvw11793/mac#:~:text=Delete%20a%20page%20from%20a,or%20choose%20Edit%20%3E%20Delete).}{Preview (on macOS)}, \href{https://www.adobe.com/acrobat/how-to/delete-pages-from-pdf.html#:~:text=Choose%20%E2%80%9CTools%E2%80%9D%20%3E%20%E2%80%9COrganize,or%20pages%20from%20the%20file.}{Adobe Acrobat} (on all OSs), as well as \href{https://superuser.com/questions/517986/is-it-possible-to-delete-some-pages-of-a-pdf-document}{command line tools}.
\clearpage
{
    \small
    \bibliographystyle{ieeenat_fullname}
    \bibliography{main}
}


\end{document}